\documentclass[conference]{IEEEtran}
\IEEEoverridecommandlockouts

\usepackage{cite}
\usepackage{amsmath,amssymb,amsfonts}
\usepackage{algorithmic}
\usepackage{graphicx}
\usepackage{textcomp}
\usepackage{xcolor}
\usepackage{subfig}

\usepackage{float}
\usepackage{tablefootnote}

\usepackage[linesnumbered,ruled]{algorithm2e}
\usepackage{multirow, booktabs}

\usepackage[pagebackref,breaklinks,colorlinks]{hyperref}
\usepackage[capitalize]{cleveref}

\usepackage{amsmath,amsfonts,bm,amsthm}

\def\eqref#1{equation~\ref{#1}}

\def\1{\bm{1}}

\def\rc{{\textnormal{c}}}

\def\rn{{\textnormal{n}}}

\def\rp{{\textnormal{p}}}

\def\ru{{\textnormal{u}}}

\def\rmM{{\mathbf{M}}}

\def\vf{{\bm{f}}}

\def\vs{{\bm{s}}}

\def\vx{{\bm{x}}}

\DeclareMathAlphabet{\mathsfit}{\encodingdefault}{\sfdefault}{m}{sl}
\SetMathAlphabet{\mathsfit}{bold}{\encodingdefault}{\sfdefault}{bx}{n}

\newcommand{\R}{\mathbb{R}}

\newcommand{\KL}{D_{\mathrm{KL}}}

\def\BibTeX{{\rm B\kern-.05em{\sc i\kern-.025em b}\kern-.08em
    T\kern-.1667em\lower.7ex\hbox{E}\kern-.125emX}}

\DeclareMathOperator{\Mix}{Mix}
\DeclareMathOperator{\Mixup}{Mixup}

\DeclareMathOperator{\PatchMix}{PatchMix}
\newtheorem{define}{Definition}
\begin{document}

\title{Learning from Ambiguous Data with Hard Labels
\thanks{$\dagger$: Correspondence to  $\langle$zekexie@hkust-gz.edu.cn$\rangle$. $*$: Equal Contributions.}
}

\author{
\IEEEauthorblockN{Zeke Xie* $\dagger$}
\IEEEauthorblockA{
\textit{HKUST(GZ)}\\
Guangzhou, China}
\and
\IEEEauthorblockN{Zheng He*}
\IEEEauthorblockA{
\textit{University of British Columbia}\\
Vancouver, Canada}
\and
\IEEEauthorblockN{Nan Lu*}
\IEEEauthorblockA{
\textit{University of Tübingen}\\
Tübingen, Germany}
\and
\IEEEauthorblockN{Lichen Bai}
\IEEEauthorblockA{
\textit{HKUST(GZ)}\\
Guangzhou, China}
\and
\IEEEauthorblockN{Bao Li}
\IEEEauthorblockA{
\textit{HKUST(GZ)}\\
Guangzhou, China}
\and
\IEEEauthorblockN{Shuo Yang}
\IEEEauthorblockA{
\textit{Harbin Institute of Technology}\\
Shenzhen, China}
\and
\IEEEauthorblockN{Mingming Sun}
\IEEEauthorblockA{
\textit{Beijing Institute of Mathematical Sciences and Applications}\\
Beijing, China}
\and
\IEEEauthorblockN{Ping Li}
\IEEEauthorblockA{
\textit{VecML Inc.}\\
Seattle, Washington, U.S.A.}
}

\maketitle
\begin{abstract}
Real-world data often contains intrinsic ambiguity that the common single-hard-label annotation paradigm ignores. Standard training using ambiguous data with these hard labels may produce overly confident models and thus leading to poor generalization. In this paper, we propose a novel framework called Quantized Label Learning (QLL) to alleviate this issue. First, we formulate QLL as learning from (very) ambiguous data with hard labels: ideally, each ambiguous instance should be associated with a ground-truth soft-label distribution describing its corresponding probabilistic weight in each class, however, this is usually not accessible; in practice, we can only observe a quantized label, i.e., a hard label sampled (quantized) from the corresponding ground-truth soft-label distribution, of each instance, which can be seen as a biased approximation of the ground-truth soft-label. Second, we propose a Class-wise Positive-Unlabeled (CPU) risk estimator that allows us to train accurate classifiers from only ambiguous data with quantized labels. Third, to simulate ambiguous datasets with quantized labels in the real world, we design a mixing-based ambiguous data generation procedure for empirical evaluation. Experiments demonstrate that our CPU method can significantly improve model generalization performance and outperform the baselines.
\end{abstract}
\section{Introduction}
\label{sec:intro}

Deep Neural Networks~(DNNs) have gained popularity in a wide range of applications. The remarkable success of DNNs often relies on the availability of high-quality datasets \cite{roh2019survey}. However, the acquisition of a large amount of well-annotated unambiguous data could be very expensive and sometimes even inaccessible \cite{liCrowdsourced}. 
In common data annotation paradigms, many tasks usually take a rather naive assumption on the ground-truth label distribution, i.e., each instance is associated with one correct hard label \cite{zhang2016learning}. 
This single-hard-label assumption seems reasonable for unambiguous data, but may result in biased labeling when an instance has high ambiguity and causes human uncertainty. \cite{yan2014learning, veit2017learning, shankar2020evaluating,wei2021learning,northcutt2021pervasive}. 

Standard training from ambiguous data with such biased hard labels induces overconfident predictions \cite{gao2017deep, vyas2020learning, peterson2019human, yun2021relabeling, xie2021artificial, xie2021positive, grossmann2022beyond, he2022sparse}.
A related line of research on biased labels is \emph{learning with noisy labels} \cite{han2020survey}, which generally studies unambiguous instances with randomly flipped noisy labels. Obviously, this classical label noise setting cannot reflect label bias from data ambiguity and human uncertainty. Therefore, a formal formulation and a novel method for learning from ambiguous data with hard labels are very much needed to address real-world challenges. 

In this work, we mainly made three contributions. 
\textbf{First}, we formally formulate the novel Quantized Label Learning (QLL) setting, where the training input distribution can be seen as an ambiguous version of the test input distribution, and the hard label modeled by the \textit{label quantization hypothesis} can be seen as a biased approximation of the ground-truth soft label. 
\textbf{Second}, to solve the QLL problem, we treat it as a soft-label and multi-class variant of Positive-Unlabeled (PU) classification task and derive a Class-wise Positive-Unlabeled (CPU) risk estimator that prevents the model from overfitting the biased hard labels.
\textbf{Third}, to provide benchmark datasets for the novel problem setting, we design a mixing-based method for preparing ambiguous datasets with hard labels and empirically evaluate the efficacy of our approach. Extensive experiments support that our CPU method can significantly improve generalization and outperform the baseline methods.
\section{Quantized Label Learning}
\label{sec:formulation}
In this section, we formulate quantized label learning as a generalized noisy data setting.

\textbf{Notations.} We denote $\mathcal{X}\subset\mathbb{R}^d$ as the instance space and $\mathcal{Y}:=[c]$ as the hard label space, where $[c]:=\{1,\dots,c\}$ and $c>2$.
In the standard classification setting, each instance $\vx_i \in\mathcal{X}$ is assigned with a hard label $y_i\in\mathcal{Y}$, and the training dataset is given by $\mathcal{D}=\{(\vx_i, y_i)\}_{i=1}^N$ where each example $(\vx_i, y_i)$ is independently and identically sampled from the joint distribution with
density $p(\vx,y)$. However, as real-world instances are ambiguous more or less, the soft-label distribution $\vs_i \in \R^c$ is considered to be the optimal annotation for an ambiguous instance $\vx_i$, as it characterizes the probability that each class label describes the instance separately. Moreover, the entropy of a soft-label distribution numerically measures the instance ambiguity.

We call the process of mapping the label distribution to a single hard label \textit{label quantization} and propose the \textit{label quantization hypothesis}: the human-annotated label of an instance is sampled from its associated ground-truth soft-label distribution. Following this, we formally define a \emph{quantized label} in Definition~\ref{df:quantizedlabel}.

\begin{define}[Quantized Label]
\label{df:quantizedlabel}
Suppose the soft-label distribution $\vs$ describes the instance $\vx$. Then, the quantized label $y$ is a hard label sampled according to the probability
\begin{equation}
P(y=k|\vx)=\frac{s_k}{\sum_{j=1}^c s_{j}},
\end{equation}
for each $k$ in $\{1,2,\dots, c\}$.
\end{define}

In QLL, we are given only ambiguous training data with quantized labels, and our goal is to learn a multi-class classifier that minimizes the (expected) classification risk:
\begin{align*}
    R(\vf)=\mathbb{E}_{(\vx,y)\sim p(\vx,y)}[\ell(\vf(\vx,y))].
\end{align*}
where $\mathbb{E}$ denotes the expectation, and $\ell: \mathcal{X}\times\mathcal{Y}\to\mathbb{R}_+$ denotes the loss function. Compared to a supervised learning setting, where unambiguous instance and hard label pairs $(\vx_i, y_i)$ drawn from $p(\vx,y)$ are available, QLL is very challenging.

\section{Class-wise Positive-Unlabeled Risk}

In this section, we first review standard Positive-Unlabeled~(PU) learning and then derive the Class-wise PU~(CPU) risk estimator for solving the QLL problem.

\begin{figure*}[t]
    \centering
    \includegraphics[width=0.80\textwidth]{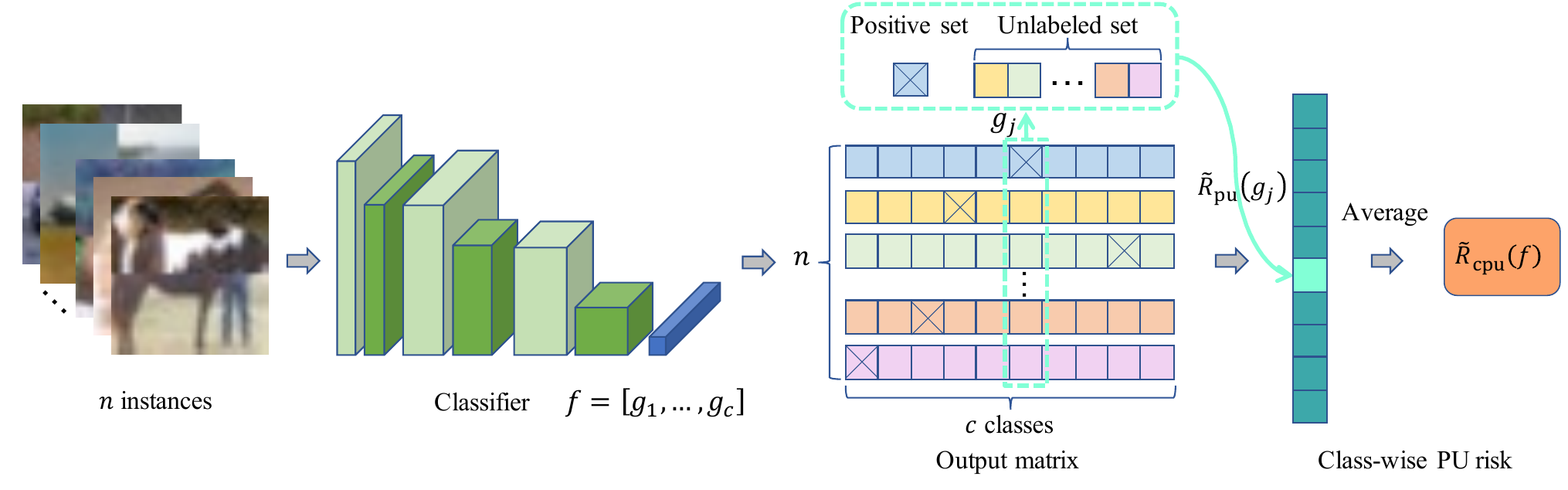}
    \vspace{-0.15in}
    \caption{Illustration of the CPU risk estimator. For a classification task with $c$ classes, we train a classifier to predict the positive/negative label for each class independently. The CPU risk is obtained by averaging the binary PU risks of $c$ classes. }
    \label{fig:cpu_algo}
\end{figure*}

\textbf{Motivation} We first consider a binary classification task to distinguish whether an instance contains information of one category or not. For example, an instance labeled as ``dog'' (quantized label) shows an observable positive relation with the ``dog'' category, but may also have features coincide with the ``wolf'' category. Treating other categories like ``wolf'' as unlabeled may help mitigate the overfitting of the quantized label ``dog''. Such a binary classification formulation with quantized labels coincides with standard PU learning setting \cite{du2014analysis, du2015convex,kiryo2017positive}, which trains binary classifiers with only a limited number of observable positive data and a large number of unlabeled data.
Motivated by this, we treat QLL as a multi-class variant of PU classification, and propose a PU approach to handle the positive and unlabeled relationships between instances and categories as described.

\textbf{PU risk.}
We first review standard hard-label PU classification before generalizing it to the soft-label and multi-class form. Let $p_{\rp}(\vx)$, $p_{\rn}(\vx)$, and $p_{\mathrm{pu}}(\vx)$ be the probability density functions of the positive, negative, and unlabeled distributions respectively, where $p_{\rp}(\vx)=p(\vx|\dot{y}=+1), p_{\rn}(\vx)=p(\vx|\dot{y}=-1)$ and $\dot{y}$ is a binary hard label. Let $\pi_{\rp} = P(\dot{y}=+1)$ and $\pi_{\rn} = P(\dot{y}=-1) = 1-\pi_{\rp}$ be the class-prior probability. Then we naturally have $p_{\mathrm{pu}}(\vx)=\pi_{\rp} p_{\rp}(\vx) + \pi_{\rn} p_{\rn}$. 

Let $g: \mathcal{X} \to \mathbb{R}$ be a binary classifier, and $l$ be the loss function. In PU learning \cite{niu2016theoretical}, the total risk is expressed in an equivalent form that can be estimated from only P data and U data, by assuming the sets of positive and unlabeled data are sampled independently from $p_{\rp}(\vx)$ and $p_{\mathrm{pu}}(\vx)$ as P data $\mathcal{X}_{\rp}=\{\vx_i^\rp\}_{i=1}^{n_{\rp}} \sim p_{\rp}(\vx)$ and U data $\mathcal{X}_{\ru}=\{\vx_i^\ru\}_{i=1}^{n_{\ru}} \sim p_{\mathrm{pu}}(\vx)$:
\begin{equation}
R_{\rp\ru}(g) = \pi_{\rp} R_{\rp}^+(g) + R_{\ru}^-(g) - \pi_{\rp} R_{\rp}^-(g), 
\label{eq:pu_risk}
\end{equation}
where $R_{\ru}^-(g)= \mathbb{E}_{\vx \sim p_{\mathrm{pu}}(\vx)}[l(g(\mathcal{X}),-1)]$ and $R_{\rp}^-(g)= \mathbb{E}_{\vx \sim p_{\rp}(\vx)}[l(g(\mathcal{X}),-1)]$. Recent work~\cite{kiryo2017positive} further proposed a non-negative PU (nnPU) risk estimator that improves the performance of the above PU risk \cref{eq:pu_risk} in the finite-sample case:
\begin{equation}
\widehat{R}_{\rp\ru}(g) = \pi_{\rp} \widehat{R}_{\rp}^+(g) + \max\{\widehat{R}_{\ru}^-(g) - \pi_{\rp} \widehat{R}_{\rp}^-(g), 0\}, 
\label{eq:nnpu}
\end{equation}
where the empirical risks are approximated by $\widehat{R}_{\rp}^+(g)=\frac{1}{n_{\rp}}\sum_{i=1}^{n_{\rp}}l(g(\vx_i^\rp), +1)$, $\widehat{R}_{\ru}^-(g)=\frac{1}{n_{\ru}}\sum_{i=1}^{n_{\ru}}l(g(\vx^\ru_i), -1)$, and $\widehat{R}_{\rp}^-(g)=\frac{1}{n_{\rp}}\sum_{i=1}^{n_{\rp}}l(g(\vx_i^\rp), -1)$.

\textbf{CPU risk.} 
In this part, we need to generalize \cref{eq:nnpu} into a class-wise form for handling multiple classes in QLL. The classifier is trained to predict the positive/negative label for each class independently given $c$ classes. A classifier $f(\vx)$ is assumed to take the following form: $f(\vx)=[g_1(\vx), g_2(\vx), \dots, g_c(\vx)]$, where $g_j(\vx)$ is a binary classifier for the class $j$ versus the rest. The binary classifier of class $j$ is denoted as $g_j^\rp(x)$ when the class $j$ is positive, and $g_j^\ru(x)$ when the class $j$ is unlabeled.

The nnPU risk estimators are calculated class-wisely by
\begin{equation}
\widehat{R}_{\rp\ru}(g_j)=\pi_{\rp} \widehat{R}_{\rp}^+(g_j) + \max\{\widehat{R}_{\ru}^-(g_j) - \pi_{\rp} \widehat{R}_{\rp}^-(g_j), 0\},
\label{eq:-cpu}
\end{equation}
where risks are estimated by separating all data into P  and U data class-wisely that 
$\widehat{R}^+_{\rp}(g_{j}) = \frac{1}{n^\rp_j}\sum_{i=1}^{n^\rp_j}l(g_j(\vx_i^\rp), +1)$, $\widehat{R}_{\ru}^-(g_j)=\frac{1}{n^\ru_j}\sum_{i=1}^{n^\ru_j}l(g_j(\vx_i^\ru), -1)$,
and $\widehat{R}_{\rp}^-(g_j) =\frac{1}{n_j^\rp}\sum_{i=1}^{n_j^\rp}l(g_j(\vx_i^\rp), -1)$, for $j \in \{1,2, \dots, c\}$.

In principle, the class prior can be prior knowledge or estimated from the observable datasets \cite{menon2015learning,jain2016estimating,christoffel2016class,yao2021rethinking}. Due to positive-negative class imbalance, in practice, we may regard the first prior $\pi_{\rp}^{(1)}$ in the positive risk term and the second prior $\pi_{\rp}^{(2)}$ in the negative risk term as two values. 
In this way, we rewrite the class-wise risk estimator as
\begin{equation}
\widetilde{R}_{\rp\ru}(g_j)=\pi_{\rp}^{(1)} \widehat{R}_{\rp}^+(g_j) + \max\{\widehat{R}_{\rn}^-(g_j) - \pi_{\rp}^{(2)} \widehat{R}_{\rp}^-(g_j), 0\}.
\label{eq:cpu-prac}
\end{equation}
Choosing the class priors is not a burden since the performance of the CPU is robust to $\pi_{\rp}^{(1)}$ and $\pi_{\rp}^{(1)}$. 

Finally, the CPU risk estimator~can~be~written~as $\widetilde{R}_{\rc\rp\ru}(f) = \frac{1}{c}\sum_{j=1}^c \widetilde{R}_{\rp\ru}(g_j)$. We illustrate the 
proposed method in \cref{fig:cpu_algo} and present the pseudocode in Algo.~\ref{algo:cpu}. We note that the inner loop can be calculated efficiently with element-wise product via modern deep learning libraries. 

\begin{algorithm}[t]
\SetKwInput{KwInput}{Input}                
\SetKwInput{KwOutput}{Output}  
  \While{no stopping criterion has been met}
  { Sample one-minibatch data and compute the model outputs $f=[g_1, g_2, \dots, g_c]$\;
     \For{$j=1$ \KwTo $c$}
     {
        \eIf{$\widehat{R}_{\ru}^-(g_j) - \pi_{\rp}^{(2)} \widehat{R}_{\rp}^-(g_j) \ge 0$}
        {
            Set $\widetilde{R}_{\rp\ru}(g_j)=\pi_{\rp}^{(1)} \widehat{R}_{\rp}^+(g_j) + \widehat{R}_{\ru}^-(g_j) - \pi_{\rp}^{(2)} \widehat{R}_{\rp}^-(g_j)$\;
        }
        {
            Set $\widetilde{R}_{\rp\ru}(g_j)=- \widehat{R}_{\ru}^-(g_j) + \pi_{\rp}^{(2)} \widehat{R}_{\rp}^-(g_j)$\;
        }
     }
    Compute the gradient of $\widetilde{R}_{\rc\rp\ru}(f) = \frac{1}{c} \sum_{j=1}^{c} \widetilde{R}_{\rp\ru}(g_j)$\;
    Update the model parameters by a optimizer;
  }
\caption{CPU Risk Minimization}
\label{algo:cpu}
\end{algorithm}
\vspace{-0.1cm}

\textbf{The choice of the loss function $l$.} As for the choice of loss function $l$, a number of statistical distance/similarity measures may serve as alternatives, such as Cross Entropy (CE), Mean Absolute Errors (MAE), Mean Squared Errors (MSE), and Kullback–Leibler (KL) divergence. Different loss functions enjoy different advantages and disadvantages in the PU framework \cite{kiryo2017positive}. In this work, we used a stochastic variant of Jensen-Shannon (JS) divergence, named Stochastic JS (SJS) divergence, as 
$D_{\rm SJS}(P\|Q)=\alpha \KL(P\|M) + (1-\alpha)\KL(Q\|M),$
where $M = \alpha P + (1-\alpha) Q$. We note that $\alpha \in (0,0.5]$ obeys a Beta distribution, $\mathrm{Beta}(0.5,0.5)$, and is randomly sampled per iteration. SJS is an interpolated generalization of KL divergence (with $\alpha \to 0$ ) and JS divergence (with $\alpha=\frac{1}{2}$). Following \cite{englesson2021generalized}, we recommend to use the scaled form $\hat{D}_{\rm SJS}(P\|Q) = -1/[(1-\alpha)\log(1-\alpha)] D_{\rm SJS}(P\|Q)$ in practice to make the loss magnitude relatively robust to $\alpha$ during training.

\section{Ambiguous Data Benchmark}
\label{sec:quantized-label-generation}

In this section, we design a mix-based method to controllably generate ambiguous data with quantized labels, which may serve as the benchmark datasets for empirical analysis.
Several methods have been proposed to generate soft labels, such as Mixup \cite{zhang2018mixup} and CutMix \cite{yun2019cutmix}. We generalize the mixing-based strategy to generate synthetic ambiguous data and soft labels with further label quantization. We call this procedure \textit{Ambiguous Data Generation}.

We denote the mix operation as $\Mix(\cdot)$ for mixing $m$ instances together. If we select $m$ instances $\{({\vx}_i, {y}_i)\}_{i=1}^m$, then the mixing process can be written as 
$\tilde{\vx} = \Mix(\vx_1, \dots, \vx_m; \boldsymbol{\lambda})$ and $\tilde{\vs} = \sum_{i=1}^m\lambda_{i}\vs_i$,
where $\boldsymbol{\lambda}=(\lambda_{1}, \dots \lambda_{m})^T$ is the mixing weight vector with the constrain $\sum_{i=1}^m \lambda_{i}=1$
and $\vs_{i}$ is the one-hot encoding of $y_i$. 
Thus, we can generate ambiguous instances with unclear features or local features obtained from different categories. We introduce two mix-based strategies as two examples, i.e., the Mixup-based generation and the Patchmix-based generation. 

By linear interpolation of instances, Mixup generates intrinsic ambiguous data which confuses annotators, especially for localization. In this case, the Mixup operation is expressed as 
$\Mixup(\vx_{1}, \dots, \vx_{m}; \boldsymbol{\lambda})= \sum_{i=1}^m \lambda_{i} \vx_{i}$.
In the Mixup-based strategy, we let $r \boldsymbol{\lambda}$ obey a $m$-category multinominal distribution, namely, $r\boldsymbol{\lambda} \sim {\rm MultiNom}(1/m,\dots,1/m; r)$. We note that $r$ is the hyperparameter that controls the number of trials of the multinominal distribution
and $\lambda_i$ indicates the frequency that the outcome $i$ occurs in $r$ trials. Similarly, we generalize the original CutMix as a novel PatchMix, expressed as
$\PatchMix(\vx_1, \dots, \vx_m; \boldsymbol{\lambda})= \sum_{i=1}^m \rmM_i \odot \vx_i$,
where each binary mask $\rmM_i$ for $\vx_i$ indicates which patches to drop out and fill in from a candidate input with the constrain $\sum_{i=1}^m \rmM_i = \boldsymbol{1}$, where $\boldsymbol{1}$ is an all-ones matrix.

\begin{table*}[h]
\caption{Best test accuracies $(\%)$ of the baselines and CPU on CIFAR-10Q and CIFAR-100Q.}
\vspace{-0.1in}
\label{tabel:results}
\centering
\small
\resizebox{0.75\textwidth}{!}{%
\setlength{\tabcolsep}{2.9mm}{
\begin{tabular}{@{}ccccccc@{}}
\toprule
\multirow{3}{*}{Dataset}  & \multicolumn{2}{c}{CIFAR-10Q} &
\multicolumn{2}{c}{CIFAR-100Q} &
\multirow{3}{*}{Mean} 
\\
\cline{2-5} 
& \multicolumn{2}{c}{Mix type} &
\multicolumn{2}{c}{Mix type} \\
  & Mixup, $m=2$ & PatchMix, $m=4$ & Mixup, $m=2$ & PatchMix, $m=4$ \\
\hline
CE        & $73.83 \pm 1.61$ & $63.30 \pm 2.09$ & $35.39 \pm 1.64$ & $22.86 \pm 0.37$ & $48.84$ \\
DMI       & $77.75 \pm 1.05$ & $63.40 \pm 3.07$ & $36.57 \pm 1.42$ & $23.05 \pm 1.57$ & $50.19$              \\
BS        & $79.24 \pm 1.25$ & $66.12 \pm 1.81$ & $33.61 \pm 1.11$ & $22.76 \pm 0.77$ & $50.88$ \\
GCE & $78.01 \pm 0.29$ & $62.42 \pm 1.62$ & $41.99 \pm 0.49$ & $23.55 \pm 0.34$ & $51.49$ \\
SCE       & $77.17 \pm 0.39$ & $59.88 \pm 1.28$ & $42.49 \pm 0.96$ & $26.69 \pm 0.74$ & $51.55$ \\
JS        & $79.19 \pm 0.63$ & $68.36 \pm 1.18$ & $46.15 \pm 0.53$ & $28.35 \pm 0.81$ & $55.51$ \\
CPU(Ours)       & \bm{$81.80 \pm 0.25$} & \bm{$72.45 \pm 0.58$} & \bm{$51.85 \pm 0.48$} & \bm{$40.25 \pm 0.38$} & \bm{$61.59$}  \\
\bottomrule 
\end{tabular}
}
}
\end{table*}

\begin{table*}[h]
\caption{Best test accuracies $(\%)$ of recent methods with complex tricks and extra hyperparameters on CIFAR-10Q/100Q.}
\vspace{-0.1in}
\label{tabel:results-adv}
\centering
\small
\resizebox{0.75\textwidth}{!}{%
\setlength{\tabcolsep}{2.6mm}{
\begin{tabular}{@{}ccccccc@{}}
\toprule
\multirow{3}{*}{Dataset}  & \multicolumn{2}{c}{CIFAR-10Q} &
\multicolumn{2}{c}{CIFAR-100Q} &
\multirow{3}{*}{Mean} \\
\cline{2-5} 
& \multicolumn{2}{c}{Mix type} &
\multicolumn{2}{c}{Mix type} \\
  & Mixup, $m=2$ & PatchMix, $m=4$ & Mixup, $m=2$ & PatchMix, $m=4$ \\
\hline
Co-teaching+ & $79.36 \pm 0.23$ & $64.66 \pm 0.68$ & $42.02 \pm 0.27$ & $24.11 \pm 0.61$ & $52.54$ \\
PES & $78.63 \pm 0.37$ & $62.46 \pm 0.93$ & $43.79 \pm 1.78$ & $29.45 \pm 1.06$ & $53.58$ \\
Co-teaching  & $80.68 \pm 0.26$ & $67.56 \pm 0.49$ & $46.74 \pm 0.12$ & $29.46 \pm 0.37$ & $56.11$ \\
ELR & $\bm{82.84 \pm 0.30}$ & $71.22 \pm 0.78$ & $49.71 \pm 0.48$ & $31.25 \pm 0.54$ & $58.76$ \\
CDR & $82.12 \pm 0.21$ & $70.68 \pm 0.79$ & $48.22 \pm 0.51$ & $35.52\pm 0.55$ & $59.13$ \\
PES (Semi) & $82.32 \pm 0.42$ & $69.96 \pm 0.52$ & $53.47 \pm 0.31$ & $36.04 \pm 1.95$ & $60.45$ \\
DivideMix & $82.53 \pm 0.44$ & $70.86 \pm 0.43$ & $54.30 \pm 0.11$ & $35.17 \pm 0.51$ & $60.72$ \\
CPU (Vanilla) & $81.80 \pm 0.25$ & $72.45 \pm 0.58$ & $51.85 \pm 0.48$ & $40.25 \pm 0.38$ & $61.59$ \\
CPU (Semi) & {$82.68 \pm 0.79$} & \bm{$74.45 \pm 0.36$} & \bm{$56.31 \pm 0.72$} & \bm{$45.70 \pm 0.59$} & \bm{$64.79$}\\
\bottomrule 
\end{tabular}
}
}
\end{table*}

\begin{table*}[h]
\caption{An ablation study of the proposed method with various loss configurations.}
\vspace{-0.1in}
\label{tabel:ablation}
\centering
\resizebox{0.8\textwidth}{!}{%
\begin{tabular}{@{}ccccccc@{}}
\toprule
\multirow{3}{*}{Dataset}  & \multicolumn{2}{c}{CIFAR-10Q} &
\multicolumn{2}{c}{CIFAR-100Q} &
\multirow{3}{*}{Mean} 
\\
\cline{2-5} 
& \multicolumn{2}{c}{Mix type} &
\multicolumn{2}{c}{Mix type} \\
  & Mixup, $m=2$ & PatchMix, $m=4$ & Mixup, $m=2$ & PatchMix, $m=4$ \\
\hline
JS\textsuperscript{*}  & $78.87 \pm 0.41$ & $66.52 \pm 0.62$ & $47.17 \pm 0.34$ & $26.50 \pm 0.62$ & $54.77$ \\
SJS (w/ scaling) & $73.73 \pm 1.62$ & $64.28 \pm 1.48$ &$43.95 \pm 0.68$ & $33.34 \pm 1.21$ & $53.83$              \\
CPU-KL (ours)   & $81.26 \pm 0.54$ & $71.33 \pm 0.93$ & $\bm{49.20 \pm 0.51}$ & $\bm{41.55 \pm 0.77}$ & $60.84$ \\
CPU (ours) & $\bm{81.80 \pm 0.25}$ & $\bm{72.45 \pm 0.58}$ & $\bm{51.85 \pm 0.48}$ & $\bm{40.25 \pm 0.38}$ & \bm{$61.59$}  \\
\bottomrule 
\end{tabular}}
\end{table*}

\section{Empirical Analysis and Discussion}
\label{sec:empirical}
In this section, we conduct extensive experiments to evaluate the effectiveness of the proposed CPU risk for QLL.

\textbf{Experimental Settings.} We use CIFAR-10/100 as the standard dataset for preparing ambiguous datasets with quantized labels, namely, CIFAR-10Q and CIFAR-100Q, following a Mixup-based method with $m=2$ and a PatchMix-based method with $m=4$ proposed in \cref{sec:quantized-label-generation}. We train ResNet-18 \cite{he2016deep} via SGD on the ambiguous training dataset. We use the original test sets of CIFAR datasets for evaluation. In practice, we may choose the class-prior hyperparameters simply as $\pi_{\rp}^{(1)}=0.1$ and $\pi_{\rp}^{(2)}=\frac{m}{c}$ without fine-tuning in our work.

\textbf{1. Various loss functions/risk estimators as the baselines.} The proposed QLL problem setting does not have baseline methods.
Given that quantized labels provide erroneous supervision for learning models, we thereby mainly compare CPU with multiple label-noise-robust baselines, including
CE, Bootstrap (BS) \cite{reed2014training}, Generalized Cross Entropy (GCE) \cite{zhang2018generalized}, Symmetric Cross Entropy (SCE) \cite{wang2019symmetric}, Determinant-based Mutual Information (DMI) \cite{xu2019l_dmi}, and JS divergence \cite{englesson2021generalized}. CE is the most common training objective/risk in standard training. Bootstrap is a famous label-noise-robust method for conventional synthetic label noise. GCE is a recent label-noise-robust generalization of CE and Mean Absolute Error. SCE is the symmetric variant of CE which is also robust to label noise. These methods mainly used designed loss functions or risk estimators, which are directly comparable to CPU.

\textbf{Results.}
 \cref{tabel:results} shows the performance of different methods on CIFAR-10Q and CIFAR-100Q under different mixing-based generation methods. We observe that our proposed risk estimator achieves the best performance across all settings. The mean accuracy of our CPU risk is higher than the second best method by more than six points.

\textbf{2. Advanced techniques/tricks as the baselines.} We also evaluate several representative advanced label-noise-robust methods and their results, including Co-teaching \cite{han2018co}, Co-teaching+ \cite{yu2019does}, CDR \cite{xia2020robust}, ELR \cite{liu2020early}, DivideMix \cite{Li2020dividemix}, PES and PES (Semi) \cite{bai2021understanding}. They do not simply use novel loss functions or risk estimators like previous baselines. Instead, they combine multiple complex techniques/tricks that were recently reported in weakly supervised learning or semi-supervised learning, including confident sample selection, early memorization, and consistency regularization. It is possible to combine CPU risk with these techniques/tricks. Thus, these methods are not directly comparable with the proposed CPU method.
We evaluate several advanced label-noise-robust methods in our empirical analysis to see whether the recent emerging tricks can handle quantized labels.

\textbf{Results.} The empirical results in \cref{tabel:results-adv} show that the mean accuracy of simply optimizing the proposed CPU risk with no trick can often outperform recent complex label-noise-robust methods that used complex tricks by one to nine points, while these methods usually require additional training budget. 
When we combine the CPU risk with simply only one semi-supervised technique FixMatch \cite{sohn2020fixmatch}, the boosted CPU (Semi) method can be further improved and easily surpass advanced rivals by four points in \cref{tabel:results-adv}.

\section{Conclusion}

Real-world data usually contains intrinsic ambiguity, which cannot be properly represented by a hard label. Unfortunately, the popular data annotation paradigm tends to assign a hard label to an instance without considering data ambiguity. In this paper, we propose a novel QLL framework where accurate classifiers can be trained from very ambiguous data with quantized labels by a CPU risk estimator. Due to the lack of benchmark datasets for QLL, we further design a mix-based ambiguous data generation procedure for evaluation. The proposed method significantly outperforms conventional baselines in this novel problem setting.

\vspace{-0.1cm}
\section*{Acknowledgment}
This work was supported by Science and Technology Bureau of Nansha District under Key Field Science and Technology Plan Program No. 2024ZD002 and Guangdong Provincial Key Lab of Integrated Communication, Sensing and Computation for Ubiquitous Internet of Things (No.2023B1212010007).

\bibliographystyle{IEEEtran}
\bibliography{reference}

\newpage
\appendix

\section{Experimental Details}
\label{apx:hpparam}

\begin{table*}[t]
\caption{Choice of optimization hyperparameters for initial learning rate (Init. LR), weight decay (WD) and method-specific hyperparameters (MS) on CIFAR-Q experiments.}
\label{tabel:hyperparameters}
\resizebox{\textwidth}{!}{%
\begin{tabular}{@{}ccccccccccccc@{}}
\toprule
\multirow{3}{*}{Method} & \multicolumn{6}{c}{CIFAR-10Q} & \multicolumn{6}{c}{CIFAR-100Q} \\ 
\cmidrule(l){2-13} 
 & 
 \multicolumn{3}{c}{Mixup, $m=2$} & \multicolumn{3}{c}{PatchMix, $m=4$} & \multicolumn{3}{c}{Mixup, $m=2$} & \multicolumn{3}{c}{PatchMix, $m=4$} \\ 
\cmidrule(l){2-13}
 & Init. LR   & WD  & MS  & Init. LR    & WD   & MS   & Init. LR   & WD  & MS & Init. LR    & WD   & MS  \\ 
 \midrule
CE  &   0.1  &  1e-4  & -   &  0.1  &  1e-4  &  -  &   0.1  &  1e-4   &   -  &  0.1  &  1e-4  &  -  \\
DMI\footnotemark
& 0.1 & 1e-4 &  - &  0.1  &  1e-4  & - & 0.1 & 1e-4 & - & 0.1 & 1e-4 & - \\
BS\footnotemark
&   0.1  &  1e-4  &  0.4  &  0.1  &  1e-4  &  0.6  &   0.1  &  1e-4   &   0.8  &  0.1  &  1e-4  &  0.8  \\
GCE\footnotemark
&   0.01  &  1e-4  & (0.7, 0.5) &  0.01  &  1e-4  &  (0.7, 0.5)  &   0.01  &  1e-4   &   (0.7, 0.5)  &  0.01  &  1e-4  &  (0.7, 0.5) \\
SCE\footnotemark
&   0.1  &  5e-4  & (0.1, 1.0) &  0.1  &  5e-4  & (0.1, 1.0) &   0.1  &  5e-4   & (6.0, 0.1) &  0.1  &  5e-4  & (6.0, 0.1) \\
JS\footnotemark
&   0.1  &  1e-3  & 0.1 &  0.1  &  1e-3  & 0.1 &   0.1  &  1e-3   & 0.1 &  1  &  1e-3  & 0.2 \\
Co-teaching+\footnotemark
&   0.1  &  1e-4  & 0.1 &  0.1  &  1e-4  & 0.1 &   0.1  &  1e-4   & 0.6 &  0.1  &  1e-4  & 0.2  \\
PES\footnotemark
&   0.1  &  1e-4  & (25, 7, 5) &  0.1  &  1e-4  & (25, 7, 5) &   0.1  &  1e-4   &  (30, 7, 5)  &  0.1  &  1e-4  &  (30, 7, 5) \\
Co-teaching\footnotemark
&   0.1  &  1e-4  & 0.3 &  0.1  &  1e-4  & 0.4 &   0.1  &  1e-4   & 0.4  &  0.1  &  1e-4  &  0.4 \\
ELR\footnotemark
&   0.02  &  1e-3  & (0.9, 1) &  0.02  &  1e-3  & (0.9, 1) &   0.02 &  1e-3   & (0.9, 7) &  0.02 &  1e-3   & (0.9, 7) \\ 
CDR\footnotemark
&   0.01  &  1e-3  & 0.3  &  0.01  &  1e-3  & 0.3  &   0.01  &  1e-3   & 0.3 &  0.01  &  1e-3  & 0.3  \\ 
PES(Semi)\textsuperscript{\ref{pes}}
&   0.1  &  1e-4  & (20, 5) &  0.1  &  1e-4  & (20, 5) &   0.1  &  1e-4   & (35, 5) &  0.1  &  1e-4  & (35, 5) \\
DivideMix\footnotemark
&   0.02  &  5e-4  & (25, 0.5)  &  0.02  &  5e-4  & (25, 0.5)  &   0.02  &  5e-4   & (25, 0.5)  &  0.02  &  5e-4  & (25, 0.5) \\ 
CPU (Vanilla)  &   0.1  &  1e-4  & (0.1, 0.2) &  0.1  &  1e-4  & (0.1, 0.4) &   0.3  &  1e-4   & (0.1, 0.02) &  0.3  &  1e-4  & (0.1, 0.04) \\
CPU (Semi)  &   0.1  &  5e-4  & (0.1, 0.2, 0.5)  &  0.1  &  5e-4  & (0.1, 0.4, 0.1) &   0.3  &  5e-4   & (0.1, 0.02, 1)  &  0.3 &  5e-4  & (0.1, 0.04, 1) \\
\bottomrule
\end{tabular}
}
\end{table*}

In this section, we present the optimization and method-specific hyperparameters used in our experiments and explain the strategy for choosing these hyperparameters.

\subsection{Experimental details on CIFAR-Q datasets}
\textbf{Optimization hyperparameters.} For all the results on the CIFAR-Q datasets, we use ResNets-18 with the standard SGD optimizer and a momentum of 0.9. 
Methods with novel loss functions or risk estimators share the same optimization mechanism for fair comparison that the batch size is set as 128, and the learning rate is reduced by a factor of 0.1 at 50\% and 75\% of the total 200 epochs (except for DMI, which sets the learning rate to decay 0.1 for every 40 epochs of total 120 epochs during the first CE training and fixed at 1e-6 for another 100 epochs during the DMI stage, following the official codes). 
The optimization mechanism of methods with advanced techniques is configured according to their official implementation. For baseline methods, hyperparameters like initial learning rate and weight decay are taken from the corresponding papers. For our methods, we use a fixed weight decay of 1e-4 and choose the learning rate in $\{0.01, 0.03, 0.1, 0.3, 1\}$ to reach a good compromise between fast convergence and robustness. We set an initial learning rate of 0.1 when training on CIFAR-10Q and 0.3 when training on CIFAR-100Q.

\textbf{Method-specific hyperparameters.}
As for the method specific hyperparameters used in these experiments, we adopt the recommended values in original papers when possible, e.g., ($q, k$) for GCE, ($\alpha, \beta$) for SCE, ($T_1, T_2, T_3$) for PES, ($T_1, T_2$) for PES (Semi), ($\beta, \lambda$) for ELR and ($\lambda_{\mu}, \tau$) for DivideMix. While some hyperparameters are sensitive to noise ratio, we tune them by choosing the best from the following grid of values $\{0.1, 0.2, \dots, 0.9\}$, e.g., $\beta$ for BS, $\pi_1$ for JS, and the estimated noise ratio $\tau$ for Co-teaching, Co-teaching+, and CDR. For our method, we set $\pi_\rp^{(1)}$ fixed and $\pi_\rp^{(2)}=m/c$ in $(\pi_\rp^{(1)}, \pi_\rp^{(2)})$ for the proposed CPU, and search the best coefficient $\lambda_{\mu} \in \{0.1, 0.5, 1, 5\}$ of the unlabeled loss when combined with FixMatch to obtain the ($\pi_\rp^{(1)}, \pi_\rp^{(2)}, \lambda_{\mu}$) for CPU (Semi). The final parameters used to get the results in Section 5 are listed in  \cref{tabel:hyperparameters}. Hyperparameters that are not discussed here use the default values from the corresponding codes.

For baseline methods, we adopt codes from the official repositories, if any, or utilize the popular unofficial PyTorch implementations.
All the experiments are conducted on a server with a single Nvidia V100 GPU.

\begin{figure}[h]
\centering     
\includegraphics[width=0.99\linewidth]{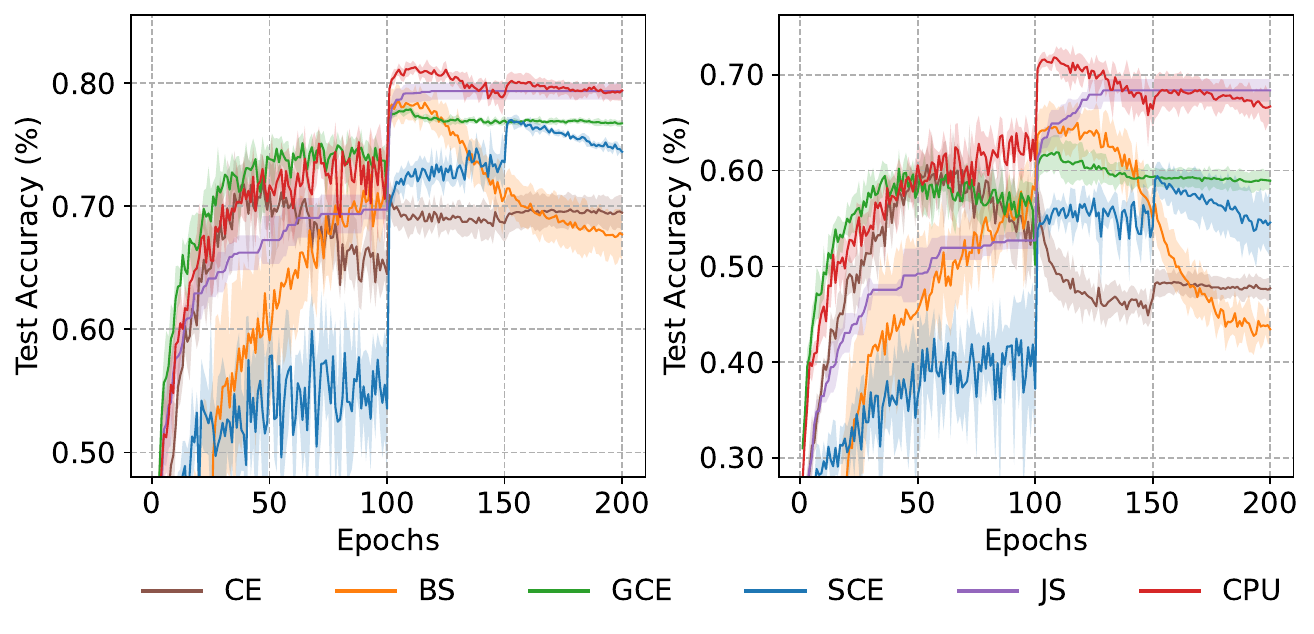}
\caption{The test curves of the baselines. Left: CIFAR-10Q, Mixup, $m=2$. Right: CIFAR-10Q, PatchMix, $m=4$.}
\label{fig:accuracy-curve-cifar10}
\end{figure}

\cref{fig:accuracy-curve-cifar10} presents the test accuracy curves of baseline methods on CIFAR10-Q. 

\subsection{Experimental details on AFHQ-Q datasets}

\begin{figure}[hhh]
	\begin{minipage}{0.32\linewidth}
		\vspace{3pt}
        \subfloat[][$y$=``cat''.]{\includegraphics[width=1.0\linewidth]{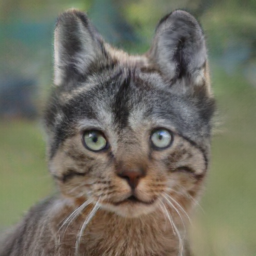}}
	\end{minipage}
        \medskip
        \begin{minipage}{0.32\linewidth}
		\vspace{3pt}
        \subfloat[][$y$=``dog''.]{\includegraphics[width=1.0\linewidth]{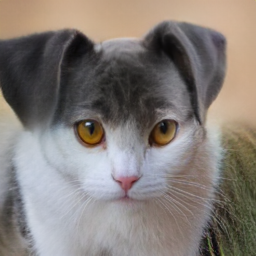}}
	\end{minipage}
        \medskip
        \begin{minipage}{0.32\linewidth}
		\vspace{3pt}
        \subfloat[][$y$=``wild''.]{\includegraphics[width=1.0\linewidth]{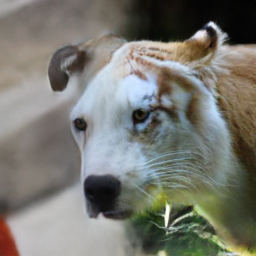}}
	\end{minipage}   
	\caption{Examples of synthetic AFHQ-Q training data with the corresponding quantized labels, generated by pre-trained StyleMapGAN. (a) Generated by ``dog'' and ``cat''. (b) Generated by ``dog'' and ``cat''. (c) Generated by ``wild'' and ``dog''. }
        \label{fig:afhq-examples}
\end{figure}

\textbf{GAN-based data generation.}
To create more realistic examples of ambiguous data, we utilize StyleMapGAN \cite{kim2021exploiting}, a GAN-based method that allows for local editing and high-quality image generation. This technique enables us to transplant selected elements of a reference image onto an original image, based on a specified mask that indicates the region to be modified. StyleMapGAN achieves this by projecting images to a latent space using an encoder, and then manipulating them locally on the latent space. By utilizing StyleMapGAN, we can mix two images and generate ambiguous examples that retain local features resembling those of their original images.

\textbf{AFHQ-Q dataset.}
The AFHQ (Animal Faces-HQ) dataset \cite{choi2020stargan} is a high-quality collection of animal face images, comprising more than 16,000 images of dogs, cats, and wildlife. Each animal category in the dataset has its own distinct visual style, and the images were carefully curated and labeled to ensure high standards.
To generate ambiguous instances of AFHQ, we utilize the pre-trained StyleMapGAN model from \cite{kim2021exploiting} and the AFHQ training data. To perform local editing on AFHQ, we randomly pair images from the training set and select half-and-half masks that divide them equally into horizontal and vertical halves (see \cref{fig:afhq-examples}). We only preserve instances that are generated by images from two different categories, resulting in 10,207 instances. The quantized labels of training data were generated as described in Section 2. The test set of AFHQ-Q consists of the original validation and test images and labels, comprising a total of 1000 images.

\textbf{Hyperparameters of AFHQ-Q experiments.}
We mainly compare our method with other robust loss on the AFHQ-Q dataset. For all these methods, we use ResNet-50 with SGD optimizer and a momentum of 0.9, and set the batch size as 128 for fair comparison. For our method, we set the learning rate decaying schedule as the CIFAR-Q experiments. And we set $\pi_p^{(2)}$ around $\frac{2}{3}$ to approximate the ambiguity in dataset. And select a larger $\pi_p^{(1)}$ values than those used in CIFAR-Q experiments because the imbalanced classification problem is less severe.  For baseline methods, we used recommended hyperparameters from the corresponding papers, but tuned some values (such as $\beta$ for BS and $\pi_1$ for JS) by choosing the best from the range of {0.1, 0.2, . . . , 0.9}. The final parameters used to get the results on AFHQ-Q are listed in \cref{tabel:hyperparameters-afhq}.

\begin{table}[h]
\caption{Choice of optimization hyperparameters for initial learning rate (Init. LR), weight decay (WD) and method-specific hyperparameters (MS) on AFHQ-Q experiments.}
\label{tabel:hyperparameters-afhq}
\centering
\small
\begin{tabular}{@{}cccc@{}}
\toprule
Method & Init. LR   & WD  & MS  \\ 
 \midrule
CE  &   0.1  &  1e-4  & -  \\
DMI & 0.1 & 1e-4 &  - \\
BS &   0.1  &  1e-4  &  0.4 \\
GCE &   0.01  &  1e-4  & (0.7, 0.5) \\
SCE &   0.1  &  1e-4  & (0.1, 1.0) \\
JS &   0.1  &  1e-4  & 0.9 \\
CPU (Vanilla)  &   0.1  &  1e-4  & (0.4, 0.6) \\
\bottomrule
\end{tabular}
\end{table}

\section{Supplementary Experimental Results}
\label{apx:supplementary-experiments}

\subsection{Robustness to the hyperparameters}

\begin{figure}[ht]
    \centering
    \includegraphics[width=0.75\linewidth]{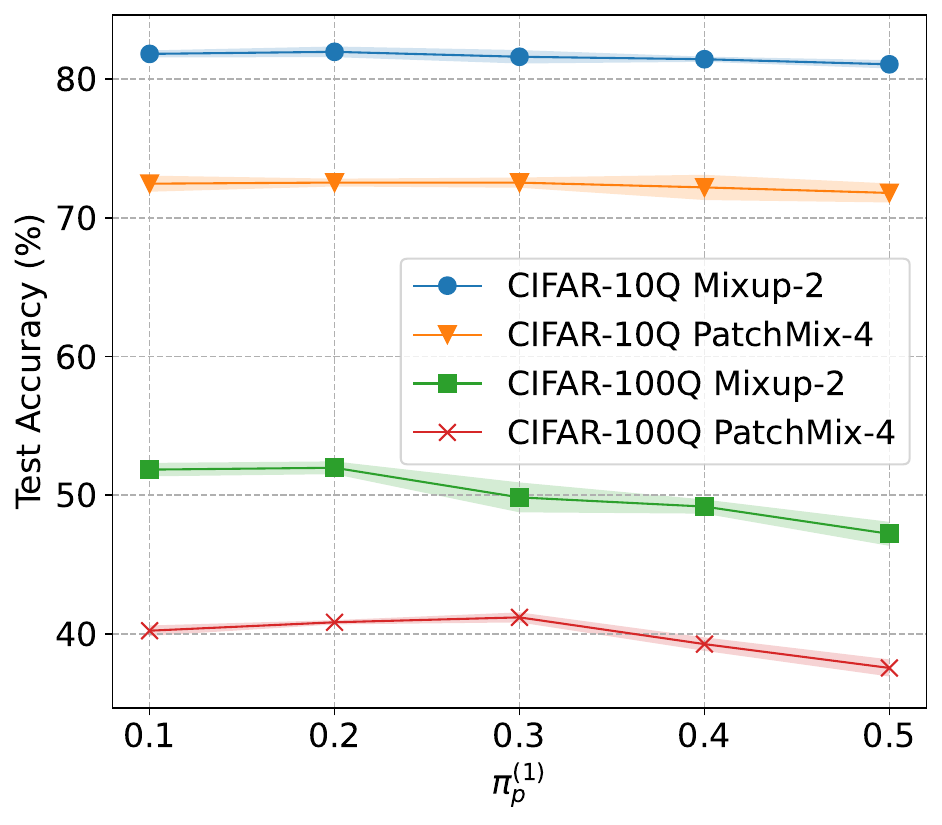}
    \caption{Illustration of robustness to different $\pi_\rp^{(1)}$. The performance of CPU is robust to various class priors.}
    \label{fig:robustness}
\end{figure}

We evaluate the performance of the CPU method under various hyperparameters $\pi_\rp^{(1)}$ and $\pi_\rp^{(2)}$. The results in Figure \ref{fig:robustness} and Table \ref{table:pi2robustness} suggest that the proposed CPU method is robust to the hyperparameters in practice. This save efforts and costs to fine-tune the new hyperparameters in CPU.

\begin{table}[h]
\caption{Robustness analysis of $\pi_\rp^{(2)}$.}
\label{table:pi2robustness}
\resizebox{\linewidth}{!}{%
\begin{tabular}{|c|ccc|ccc|}
\hline
CIFAR-10Q  & \multicolumn{3}{c|}{Mixup, k=2} & \multicolumn{3}{c|}{Patchmix, k=4} 
\\ \hline
$\pi_\rp^{(2)}$     & \multicolumn{1}{c|}{0.15} & \multicolumn{1}{c|}{0.2}  & 0.25 & \multicolumn{1}{c|}{0.35} & \multicolumn{1}{c|}{0.4}  & 0.45
\\ \hline
Best test acc. & \multicolumn{1}{c|}{$81.21$}     & \multicolumn{1}{c|}{81.80}     & $81.46$  & \multicolumn{1}{c|}{$72.69$}     & \multicolumn{1}{c|}{72.45}     &   $71.57$                    \\ \hline
CIFAR-100Q & \multicolumn{3}{c|}{Mixup, k=2} & \multicolumn{3}{c|}{Patchmix, k=4} \\ \hline
$\pi_\rp^{(2)}$      & \multicolumn{1}{c|}{0.01} & \multicolumn{1}{c|}{0.02} & 0.03 & \multicolumn{1}{c|}{0.03} & \multicolumn{1}{c|}{0.04} & 0.05                  \\ \hline
Best test acc. & \multicolumn{1}{c|}{51.54}     & \multicolumn{1}{c|}{51.85}    &   51.33   & \multicolumn{1}{c|}{39.73}     & \multicolumn{1}{c|}{40.25}     & \multicolumn{1}{c|}{39.94} \\ \hline
\end{tabular}%
}
\end{table}

Combining CPU with KL loss outperforms on challenging datasets like CIFAR-100Q generating by PatchMix, while becomes inferior on CIFAR-10Q because of overfitting. Contrarily, CPU with JS\textsuperscript{*} loss is more robust to CIFAR-10Q, but performs poorly on CIFAR-100Q because of the difficulty to converge. By adopting a stochastic variant of JS\textsuperscript{*} along with CPU, our proposed method becomes competitive across all datasets, which suggests the stochastic sampling of $\alpha \in (0, 0.5]$ during training satisfies both optimization and generalization. Moreover, in order to avoid a unreasonably small value of training loss, scaling the magnitude of CPU-SJS with respect to $\alpha$ is a must especially on CIFAR-100Q.

\footnotetext[1]{https://github.com/Newbeeer/L\_DMI}
\footnotetext[2]{https://github.com/vfdev-5/BootstrappingLoss}
\footnotetext[3]{https://github.com/AlanChou/Truncated-Loss}
\footnotetext[4]{https://github.com/HanxunH/SCELoss-Reproduce}  
\footnotetext[5]{https://github.com/ErikEnglesson/GJS\label{gjs}}  
\footnotetext[6]{https://github.com/xingruiyu/coteaching\_plus}  
\footnotetext[7]{https://github.com/tmllab/PES\label{pes}}  
\footnotetext[8]{https://github.com/bhanML/Co-teaching}  
\footnotetext[9]{https://github.com/shengliu66/ELR}  
\footnotetext[10]{https://github.com/xiaoboxia/CDR}  
\footnotetext[11]{https://github.com/LiJunnan1992/DivideMix}

\subsection{Results on AFHQ-Q dataset}

To create benchmark ambiguous datasets that look more realistic for humans, we utilize StyleMapGAN \cite{kim2021exploiting} to mix two images and obtain ambiguous examples (see Figure \ref{fig:afhq-examples}) that retain local features resembling those of their original images. In Table \ref{tabel:cpuafhq}, we compare CPU with several representative baselines, including VLE \cite{xu2020variational}, DMI, BS, Co-teaching, and Co-teaching+. We leave the details in the appendix. Obviously, the label-noise methods, Co-teaching and Co-teaching+, cannot handle such ambiguous dataset well. This suggest the significant difference between the quantized label learning problem and the noisy label learning problem.

\begin{table}[h]
\caption{Last 5 epoch average and best test accuracy $(\%)$ of the baselines and CPU on AFHQ-Q dataset.}
\vspace{-0.1in}
\label{tabel:cpuafhq}
\centering
\small
\resizebox{0.48\textwidth}{!}{%
\setlength{\tabcolsep}{2.9mm}{
\begin{tabular}{@{}cccc@{}}
\toprule
\multirow{1}{*}{Method}  & 
\multicolumn{1}{c}{Last 5 epoch average acc.} &
\multicolumn{1}{c}{Best test acc.} &
\multirow{1}{*}{Mean} 
\\
\hline
VLE       & $41.68 \pm 2.28$ & $53.16 \pm 1.02$ & $47.42$ \\
DMI       & $62.64 \pm 20.08$ & $78.66 \pm 6.94$ & $70.65$  \\
BS        & $75.156 \pm 9.35$ & $85.66 \pm 6.70$ & $80.41$ \\
Co-teaching      & $87.21 \pm 1.59$ & $89.12 \pm 0.65$ & $88.17$ \\
Co-teaching+      & $64.39 \pm 9.75$ & $70.64 \pm 3.22$ & $68.73$ \\
CPU(Ours)       & \bm{$91.01 \pm 1.10$} & \bm{$94.28 \pm 1.09$} & \bm{$92.64$} \\
\bottomrule 
\end{tabular}}
}
\end{table}

\section{Related Work}
\label{sec:related}

In this section, we discuss related studies.

\textbf{Data Ambiguity.}
Recent studies \cite{beyer2020we, yun2021relabeling, shankar2020evaluating, tsipras2020imagenet, grossmann2022beyond, Schmarje2021ADA} have shed light on an overlooked problem that many real-world datasets contain data with intrinsic ambiguity.
Annotations of ambiguous data may be incorrect or misaligned with the
ground truth \cite{tsipras2020imagenet}.
Thus, learning with ambiguous data and hard labels degrades model performance \cite{grossmann2022beyond}.
While these works mainly explored dataset relabeling, we are the first to formulate the problem of learning with ambiguous data and quantized labels.
Otani \cite{otani2020binary} also assumed the existence of ambiguous inputs and propose to reject them during training. However, their setting can only apply to the binary classification and is not as challenging as QLL, where models are trained with and only with ambiguous data. \emph{Partial label learning} \cite{cour2011learning, zhang2015solving, zhang2017disambiguation} is a related topic on ambiguous labels, which considers the multi-label description of each instance. But their problem setting assumes that only one of the multiple candidate labels is true and the intrinsic data ambiguity is not considered. 

\textbf{Noisy Labels.}
Learning with noisy labels is a related topic in the sense of considering ambiguous labels, where their training labels are corrupted versions of the ground-truth labels
\cite{han2020survey}. 
Prior studies widely adopted the instance-independent label noise, (i.e., the label flip rates are only class-dependent) \cite{reed2014training, zhang2018generalized, wang2019symmetric, han2018co, Li2020dividemix, xie2021artificial,xie2021positive}, but
real-world label errors usually come from human mistakes that are highly instance-dependent \cite{wei2021learning}. 
There are some recent techniques for synthesizing the instance-dependent label noise \cite{cheng2020learning, xia2020part}. However, label corruption along this line of research can be modeled as a label noise transition matrix, while label quantization has no label noise transition in the hard-label space. These label-noise works just assume that each instance has only one correct label and neglect the inherent ambiguity in data. This clearly distinguishes the discussions of noisy labels from quantized labels.

\textbf{Soft Labels.} 
Label distribution learning proposes that each training example is associated with a soft label, and the goal is to predict the label distribution of test data \cite{geng2016label, gao2017deep}. 
However, label distribution is hard to acquire due to high annotation costs.
In contrast, the quantized label hypothesis is more realistic for real-world datasets.
Furthermore, soft labels have been demonstrated to improve model calibration or enhance robustness \cite{muller2019does, zhang2020does, peterson2019human}. The problem of learning with quantized labels is worth greater attention, in order to overcome the performance bottleneck of learning models.


\end{document}